# ROBUST DUAL VIEW DEEP AGENT

IBRAHIM M. SOBH[1], NEVIN M. DARWISH[2]

**ABSTRACT**

Motivated by recent advance of machine learning using Deep Reinforcement Learning this paper proposes a modified architecture that produces more robust agents and speeds up the training process. Our architecture is based on Asynchronous Advantage Actor-Critic (A3C) algorithm where the total input dimensionality is halved by dividing the input into two independent streams. We use ViZDoom, 3D world software that is based on the classical first person shooter video game, Doom, as a test case. The experiments show that in comparison to single input agents, the proposed architecture succeeds to have the same playing performance and shows more robust behavior, achieving significant reduction in the number of training parameters of almost 30%.

**KEYWORDS**: Deep Reinforcement Learning.

## 1. INTRODUCTION

Deep learning has contributed to dramatic advances in scalability and performance of machine learning [1]. Deep Learning enabled breakthroughs in variety of domains, for example, image classification [2], speech recognition [3], machine translation [4], object recognition [5] and robot control [6]. Recently, deep reinforcement learning has proved to be successful in learning human-level control policies in different domains. In the computer games domain, deep reinforcement learning is shown to be effective in playing simple Atari 2600 games [7 - 9] and in defeating world class Go players [10]. In the above applications, the full knowledge of the current state of the environment is assumed. However, in real-world scenarios, states are partially observable. In such case, the agent needs to select optimal actions based on a sequence of previous states.

[1] Cairo University, Faculty of Engineering, Computer Department, imsobh@gmail.com
[2] Cairo University, Faculty of Engineering, Computer Department, ndarwish@eng.cu.edu.eg

Recurrent Neural Networks (RNN) can remember information for an arbitrarily long sequence and therefor they are effectively used for partially observable situations.

Previous methods have mainly been applied on 2D environments that are not similar to real world situations. In this paper, we tackle the task of playing a First-Person-Shooting (FPS) game in a semi realistic 3D environment. This task is much more challenging compared to most Atari 2D games. In 3D environments states are partially observable. Generally, the agent navigates into a 3D environment, shoots enemies and collects items from a first-person perspective. These conditions make the task more suitable for real-world applications.

In Doom, a classic FPS game, the player controls the agent to fight against enemies, collect objects and explore a 3D environment. The agent can partially see the environment, and take decisions accordingly. ViZDoom [11] is an open-source platform that offers a programming interface to communicate with the Doom game engine. From the interface, users can obtain the current frame of the game, take actions and receive feedback rewards. Built in bots with different capabilities can be inserted into the game. Furthermore, ViZDoom environment offers games variables including agent's health and spatial locations.

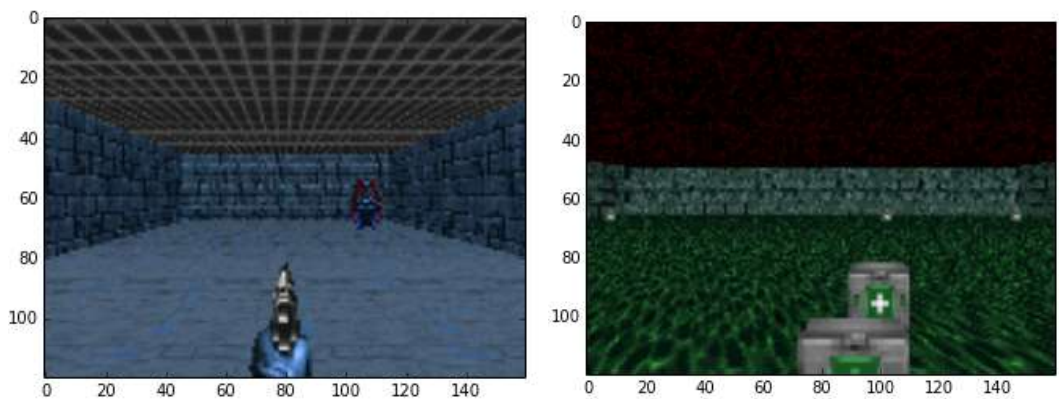

Figure 1. Left: Basic Shooting Scenario; Right: Health Gathering Scenario

In this work, we present an agent for playing Doom using only the raw pixels, as viewed on the screen, in addition to occasional feedback rewards from the simulation environment. Input dimensionality is halved by dividing the input frame into two views: Generic low resolution view, and Centered high resolution view. We evaluate our agent for two main scenarios namely; Basic shooting and Health gathering (Figure 1). We show that the proposed architecture is a promising way for robust performance in case of partial input failure and also for speeding up the training process.

## 2. BACKGROUND

For purpose of self-containment, we start by a brief summary of the Reinforcement Learning (RL), Deep Q-Networks (DQN), Deep Recurrent Q-Networks (DRQN) and Asynchronous Advantage Actor-Critic (A3C) algorithm.

### 2.1 Reinforcement Learning

Reinforcement learning deals with learning a policy for an agent interacting in an unknown environment. At each step, an agent observes the current state of the environment, then decides to take an action according to a policy. Occasionally, the agent receives a feedback signal from the environment as a reward or penalty. The goal of the agent is to find a policy that maximizes the expected discounted return *Rt*:

$$R_t = \sum_{T=t}^{\infty} \gamma^{T-t} r_T \tag{1}$$

Where the discount factor $\gamma \in [0, 1]$ determines the importance of the future rewards. Q-Learning [12] estimates state-action value function using a Q-function, which measures the value of choosing a particular action when being in a particular state. The Q-function of a given policy π is defined as the expected return from executing an action *a* in a state *s*:

$$Q^\pi(s, a) = E[R_t \mid s_t = s, \ a_t = a] \quad (2)$$

The optimal Q-function is defined as: $Q^*(s, a) = max_\pi Q^\pi(s, a)$, which verifies the Bellman optimality equation:

$$Q^*(s, a) = E[r + \gamma \, max_{a'} Q^*(s', a') \mid s, a] \quad (3)$$

## 2.2 Deep Q-Networks (DQN)

In real world applications, there are too many number of states. DQN uses a neural network parametrized by θ as an approximator for the action-value function. The goal is to find the parameters θ such that: $Q(s, a, \theta) \approx Q^*(s, a)$, where (s, a) represents the state action pair. We find θ by optimizing the loss function:

$$L_t(\theta_t) = E[(y_t - Q(s, a, \theta_t))^2] \quad (4)$$

Where $y_t = r_t + max_{a'} Q(s', a', \theta^-)$

For algorithm stability, according to [8], the parameters of the target network ($\theta^-$) are frozen for a fixed number of iterations while updating the online network parameters ($\theta_t$) using gradient descent.

It is important to use experience replay to break the possible correlation between successive training samples [13]. Therefore, experiences of an agent are stored in a "replay memory" or queue, Then, Q-learning updates are done on batches of experiences which are randomly sampled from the replay memory. Moreover, experience replay increases data efficiency because of the reuse of experience samples in multiple updates.

At every training step, the next action is generated using an ε-greedy strategy where the next action is selected randomly with probability ε , and the best known action generated by the network is selected with probability 1-ε. Usually, ε starts with 1 and decays gradually. Intuitively, the decaying process encourages exploration at the beginning of training. DQN is off-policy because states and rewards are obtained with an epsilon greedy policy, the behavior policy, which is different from the online policy being learned. DQN improvements include dueling architectures [14], prioritized experience replay [15], and double q-learning [16].

### 2.3 Deep Recurrent Q-Networks (DRQN)

Regarding FPS-3D games, the agent field of view is limited around its position and direction. The Deep Recurrent Q-Networks (DRQN) [17] are introduced to deal with partially observable environments. In DRQN a recurrent neural network, Long Short Term Memory LSTM, is used on top of the normal DQN model to enable representing the hidden state of the agent. DRQN is capable of integrating information across a sequence of frames. When trained on standard Atari 2D games and evaluated using flickering inputs, DRQN showed better performance compared to DQN .

### 2.4 Asynchronous Advantage Actor-Critic (A3C)

Deep RL algorithms based on experience replay such as DQN and DRQN have succeeded in domains such as 2D Atari games. However, experience replay uses more memory to store experience and requires off-policy learning algorithms. A different paradigm is introduced in [9] for deep reinforcement learning. Instead of using experience replay buffer, independent agents asynchronously execute on multiple parallel instances of the environment. In addition to the decorrelation of training data, the parallel learners have a stabilizing effect on the training process. Moreover, this simple setup enables a much larger spectrum of on-policy as well as off-policy reinforcement learning algorithms to be applied. The best performing method found was an asynchronous variant of the actor-critic, namely Asynchronous Advantage Actor-Critic (A3C) that exceeds the state-of-the-art on the Atari domain while training for half the time on a single multi-core CPU. A3C model shows good performance for 3D labyrinth environment exploration.

## 3. THE PROPOSED ARCHITECTURE

In general, training an agent in a partially observable 3D environment from raw frames remains an open challenge. The main goal of the proposed architecture is to train a robust agent and to speed up the training process. The proposed agent architecture is based on the network architecture from [9]. The network uses a convolutional layer with 16 filters of size 8 × 8 with stride 4, followed by a convolutional layer with 32 filters of size 4 × 4 with stride 2, followed by a fully connected layer with 256 hidden units. We add an LSTM layer of 256 units. All hidden layers are followed by Exponential Linear Units (ELU) [18]. We have two outputs represented by two heads. The policy that represents probability distribution on actions, and implemented as a softmax output with one entry per action. And the value that represents a single linear output. Experiments have a discount factor $\gamma = 0.99$ and an Adam optimizer [19] with a learning rate of 1e-4. Following [9], we use entropy regularization with a weight of 0.01 which has the effect of improving the exploration by discouraging naive convergence to suboptimal policies.

## 3.1 Dual View Input

The original A3C model uses a single input view. However, our proposed architecture uses two views. Usually, the agent sees frames of the game, each frame is 84 × 84 pixels. Instead, we split input view into a dual view as follows:

1. Generic View: 42 × 42 low resolution of the original frame.
2. Center View: 42 × 42 high resolution of the frame center.

The idea is to process each frame separately allowing the agent to have two eyes or two perspectives of the input; one is for the overall, generic view, and the other is for the focused, center view (Figure 2).

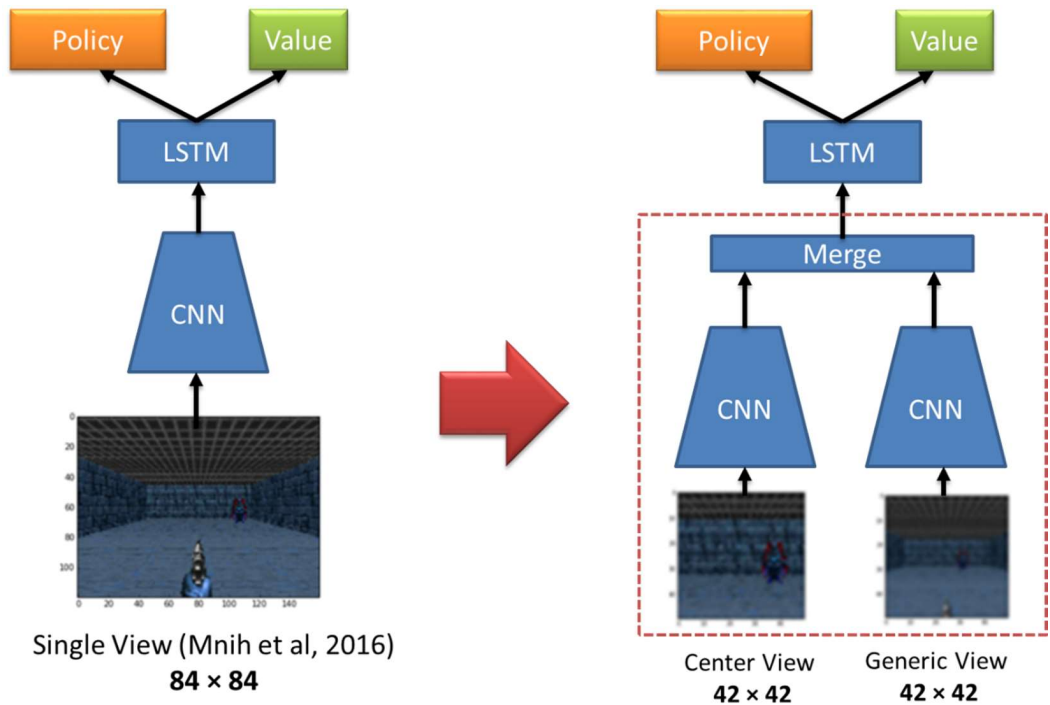

Figure 2. Left: The original architecture of [9]; Right: the proposed dual view, half input architecture

This setup has the advantage of reducing the input size by half. Intuitively, this is suitable for FPS games and a wide range of applications.

## 4. EXPERIMENTS

We use two well-known scenarios of Doom, Basic shooting and Health gathering scenarios. For network training and testing, we use TensorFlow [20], an open source software library for numerical computation using data flow graphs.

### 4.1 Basic Shooting Scenario

This simple scenario takes place in a rectangular room. The agent is located in the center of the room's longer wall. A stationary monster is located at an arbitrary position along the opposite wall. The agent can move left and right, or shoot. A single hit is enough to kill the monster. The episode ends when the monster is eliminated or after 300 frames, whatever comes first. The agent scores +100 points for killing the monster, and −1 point for each action. These scores motivate the learning agent to kill the monster as quickly as possible.

### 4.2 Health Gathering Scenario

In this scenario, the agent is located in a random spot of an open area surrounded by walls where the surface is covered by a harmful acid, which takes away the agent's health gradually. To survive, the agent needs to collect the medical kits (medikits) which are located in random places. The agent can turn left and right, or move forward. The episode length is 2100 frames or when the agent is dead, whatever comes first. The agent scores +5 points for picking a medical kit and +1 for being alive. This motivates the learning agent to pick medical kits to stay alive for the longest period possible.

**4.3   Skip Count**

As mentioned by [11], when the agent does not skip any frames, the learning is the slowest, and by using larger skip-counts, the learning is faster. On the other hand, using too large skip-counts makes the agent suffer from lack of fine-grained control as it repeats a large number of actions, which results in suboptimal final scores. Considering the two scenarios, we found that skip-counts equals to 4 provides a good balance between the learning speed and the final performance.

**5.   RESULTS**

In this section, we show the comparisons of training convergence for single and dual view agents. Then, detailed robustness tests are presented. Finally, we use saliency maps visualizations to understand how the network takes decisions given the input frames.

**5.1   Training Convergence**

We compare the training convergence behavior of single and dual view agents for both basic shooting and health gathering scenarios. In the following Figure 3 and Figure 4, the average performance of 8 parallel learning agents, each on its own environment, is plotted for single and dual view cases.

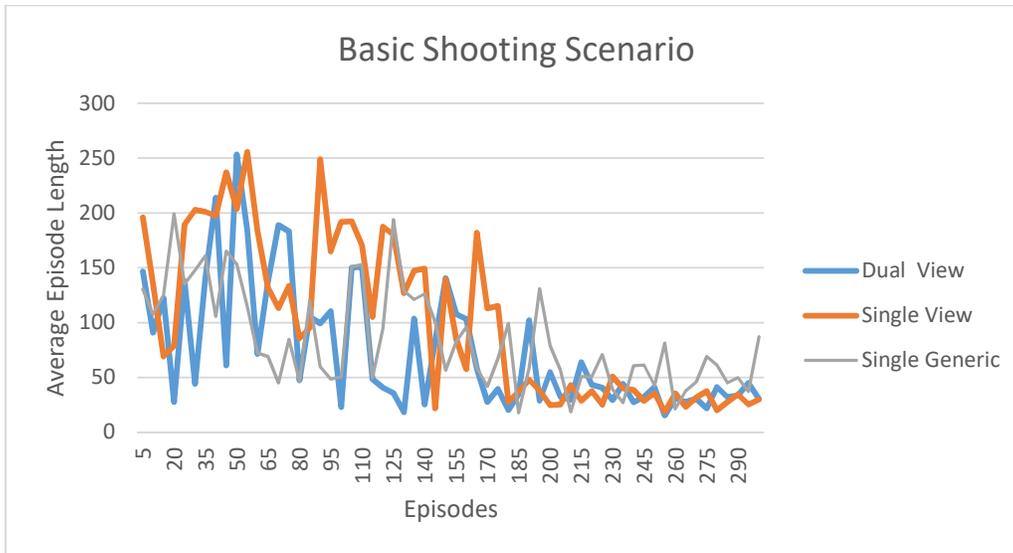

Figure 3. Basic Shooting Scenario: training convergence comparison

In the basic shooting scenario, the agent is supposed to kill the monster as fast as possible, less episode time is better. Figure 3 shows a comparison between single and dual view agents for the basic shooting scenario, in addition to single view agent using only generic view. Both, single and dual view agents have comparable convergence behavior. However, the single view agent that uses only generic view could not finish the episode at shorter time. The main reason is that high shooting accuracy cannot be easily achieved depending only on one generic low resolution view. As shown, dual view agent shows a marginal higher variance in the beginning of training compared to single view agent. However, the dual view agent has 30% less parameters than the single view agent. Similarly, Figure 4 illustrates convergence behavior for the health gathering scenario, where agent goal is to survive by collecting medical kits, where surviving for longer episode is better. In a longer and more complicated scenario of health gathering, both single and dual view agents have very comparable convergence behavior. On the other hand, the single generic view agent managed to find a good policy but not as good as the single view and dual view agents.

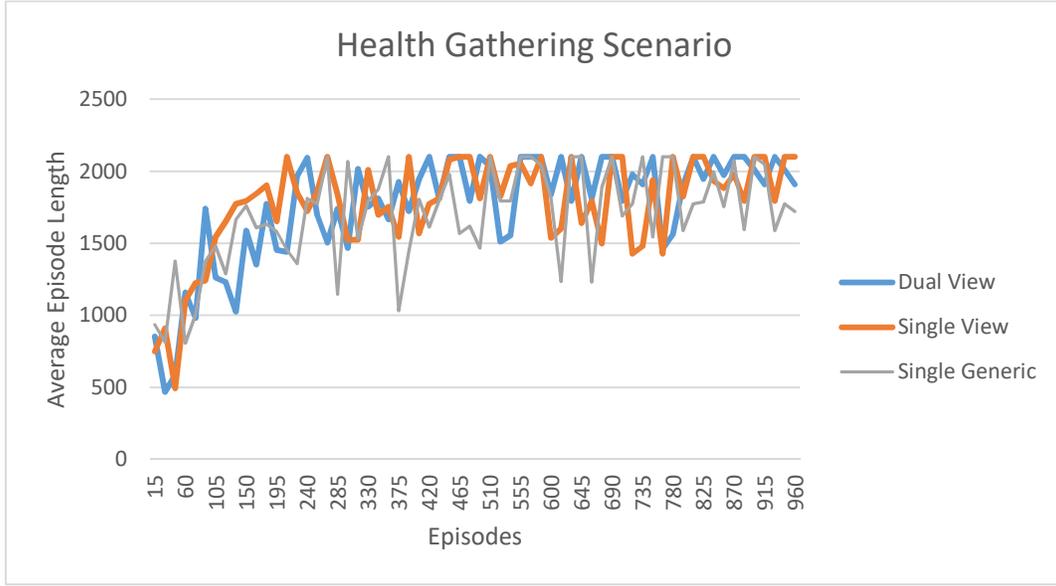

Figure 4. Health Gathering Scenario: training convergence comparison

## 5.2 Robustness

In real world applications, it is important to test an agent's behavior in case of sensory failure. Here, we address the possibility of missing input information, and how this affects the agent's performance. We test the trained agent for both scenarios, where the input view is dropped, and the corresponding frame is totally black. We define frame drop probability as $P_{drop}$. In order to measure the effect of dropped frames we define the agent's score percentage of the original score, $S_p$ as follows:

$$S_p = \frac{S_a - S_{min}}{S_{max} - S_{min}} \tag{6}$$

Where $S_a$, is the average score of the agent after applying input drop. $S_{min}$, is the minimum score of the agent. In other words, the score that agent may achieve without training. Finally, $S_{max}$, is the maximum score of the agent after training. Generally, dropping input frames causes performance degradation, where the agent tries to generalize and deal with the missing information. Intuitively, for a dual view agent, when both views are dropped at $P_{drop} = 1.0$ the input to the agent is just black frames. Accordingly, the agent acts randomly and hence $S_a = S_{min}$ and $S_p = 0$. On the other hand, when both views are dropped at $P_{drop} = 0.0$, the input to the agent is complete, and the agent performs as trained giving its maximum original score where, $S_a = S_{max}$ and $S_p = 1$. In the following experiments, we change the $P_{drop}$ for both scenarios in single and dual view cases. Table 1 shows the dual view basic shooting scenario score for different frame drop probabilities.

Table 1. Frame dropping performance: Dual view Basic Shooting Scenario

| Center View $P_{drop}$ | | | | | |
|---|---|---|---|---|---|
| 1.0 | **60.6%** | 54.7% | 40.6% | 25.3% | **0.0%** |
| 0.8 | 87.1% | **78.2%** | 52.9% | 33.5% | 13.5% |
| 0.5 | 88.8% | 80.6% | 71.2% | 35.9% | 31.2% |
| 0.2 | 95.3% | 86.5% | 70.0% | **42.9%** | 36.5% |
| 0.0 | **100.0%** | 90.6% | 72.9% | 47.6% | **41.7%** |
| | 0.0 | 0.2 | 0.5 | 0.8 | 1.0 |
| | Generic View $P_{drop}$ | | | | |

The results imply that the effect of frame drop is not the same for generic and center views. For example, when the generic view is dropped at probability of 0.8, and the center view is dropped at probability of 0.2, then $S_p = 42.9\%$, while when the situation is reversed, where the generic view is dropped at probability of 0.2, and the center view is dropped at probability of 0.8, then $S_p = 78.2\%$. Another extreme case, when the center view is totally dropped and the generic view is totally reserved, then $S_p = 60.6\%$. And when the generic view is totally dropped and the center view is totally reserved, then $S_p = 41.7\%$. These results indicate that the agent depends more on the generic view to decide what action to take. The center view still adds information and increases the $S_p$ as it enhances the shooting accuracy and provides added information when the generic view is dropped.

Regarding the case of Single view agent, it has only one input, when completely dropped, it has no other information source to make use of. On the other hand, the dual view agent has the advantage that it still has a chance to make use of the dual input, and performs better than randomly, even if one of the inputs is completely dropped.

Table 2. Frame dropping performance: Single view Basic Shooting Scenario

| 100.0% | 79.3% | 63.6% | 24.1% | 0% |
|---|---|---|---|---|
| 0.0 | 0.2 | 0.5 | 0.8 | 1.0 |
| Main View $P_{drop}$ ||||| 

Table 2 shows the frame dropping performance for the Single view agent. It is noted that, when the main view is dropped at probability of 0.2, then $S_p = 79.3\%$ which is lower than the dual view agent when the generic view is dropped at the same probability and the center view is dropped at probabilities of 0.0, 0.2 and 0.5 where the dual view agent performance is $S_p = 96.6\%, 86.5\%\ and\ 80.6\%$ respectively. This indicates that using dual view increases the robustness of the agent in case of sensory partial failure in a much better way when compared to the single view agent. Another extreme comparison is shown in Figure 5. Typically, the dual view agent is trained to make use of both views. Here, we challenge the dual view agent where one of its views is totally dropped. The first case illustrates the case where generic view is used and the center view is totally dropped. The second case gives the other extreme case where center view is used and the generic view is totally dropped. The third and final case, shows the single view high resolution agent.

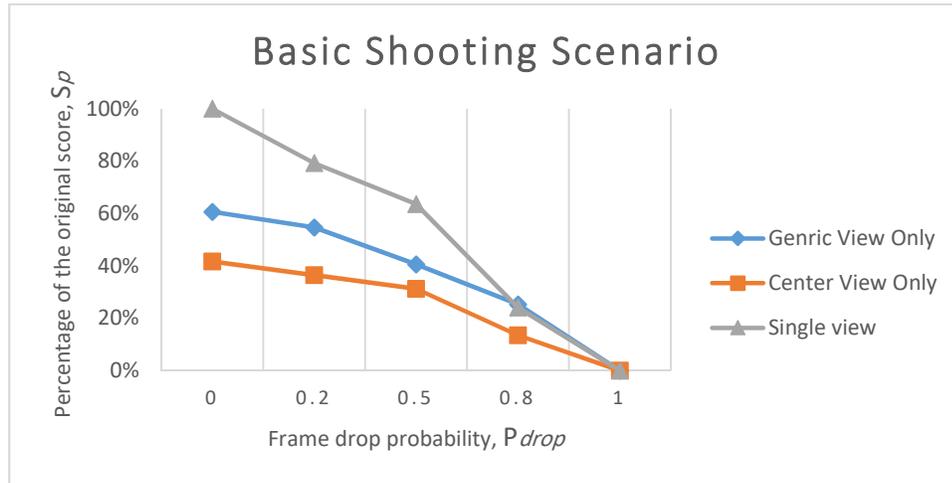

Figure 5. Basic Shooting Scenario: Comparing robustness for single view agent and dual view agent where only one view is considered

In Figure 5, it is clear that with frame drop probabilities of 0, 0.2 and 0.5, the single view agent has better performance. The main reason is that the dual view agent has no information at all from the totally dropped view. On the other hand, the dual view agent with only the center view has comparable performance to the single view agent when frame drop probability is 0.5 and above. As expected, centric view is still useful, but using it alone is challenging for the agent.

Table 3 shows the dual view health gathering scenario scores for different frame drop probabilities.

Table 3. Frame dropping performance: Dual view Health Gathering Scenario

| Center View $P_{drop}$ | | | | | |
|---|---|---|---|---|---|
| 1.0 | **72.8%** | 46.1% | 26.4% | 18.6% | **0.0%** |
| 0.8 | 71.9% | 48.4% | 23.5% | 25.5% | 2.3% |
| 0.5 | 76.2% | 51.9% | 31.3% | 26.4% | 17.7% |
| 0.2 | 84.3% | 55.9% | 31.9% | 26.7% | 19.7% |
| 0.0 | **100.0%** | 73.0% | 37.7% | 25.5% | **20.0%** |
| | 0.0 | 0.2 | 0.5 | 0.8 | 1.0 |
| | Generic View $P_{drop}$ | | | | |

Similar to the basic shooting scenario, the agent depends more on the generic view. The center view in health gathering scenario enhances agent's performance, but it has a smaller contribution when compared to the basic shooting scenario. As shown in Table 4, unlike shooting, the single view agent showed better performance for dropping probabilities of 0.2 and 0.5, compared to the dual view agent. The reason is that, shooting requires detailed information about what appears in front of the agent, while health gathering requires the generic view to observe where the medical kits are located and act accordingly.

Table 4. Frame dropping performance: Single view Health Gathering Scenario

| 100.0% | **82.9%** | **49.0%** | 25.3% | 0% |
|---|---|---|---|---|
| 0.0 | 0.2 | 0.5 | 0.8 | 1.0 |
| Main View $P_{drop}$ | | | | |

Similar to basic shooting scenario, Figure 6 shows how performance is decreased when increasing the frame drop probability. However, the performance of the center view by itself drops much faster than the case of the basic shooting scenario.

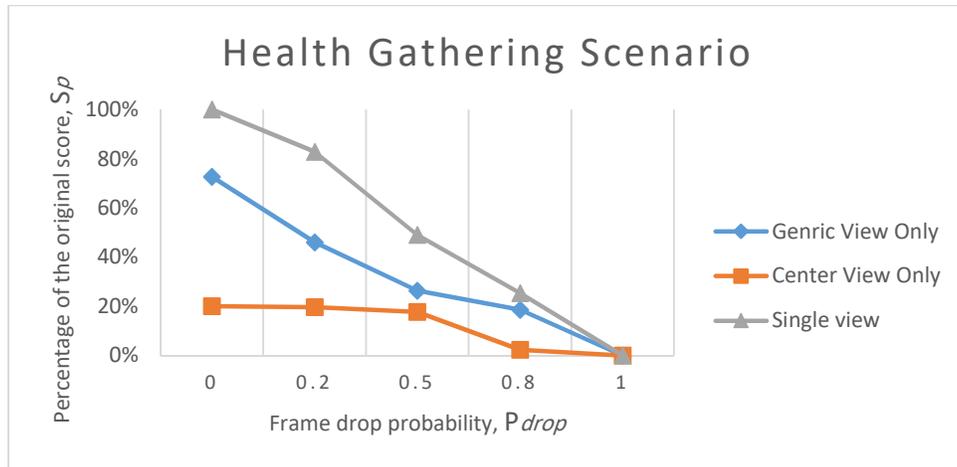

Figure 6. Health Gathering Scenario: Comparing robustness for the single view agent and the dual view agent where only one view is considered

### 5.3 Saliency Maps

We use saliency maps [21] to visualize the salient parts of the input frames as considered by both value and policy networks. We compute the absolute value of the Jacobian of value and policy with respect to the input frames. Saliency maps have the same spatial dimensionality as the input frames and can be visualized to indicate where the network pays more attention. Saliency maps shown in Figures 7 and 8, are indicating that the network conducts a reasonable learning.

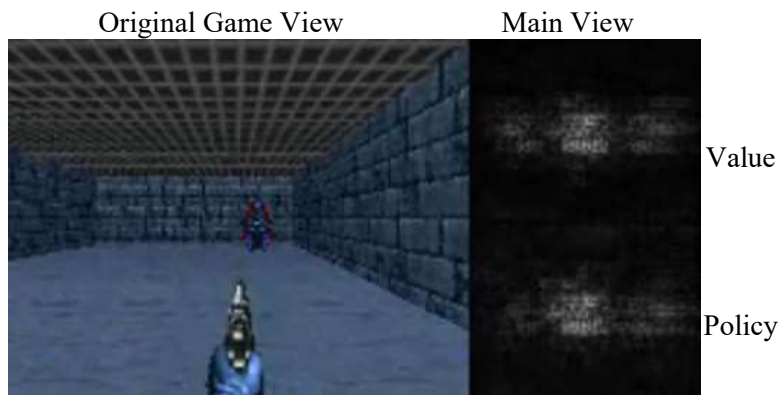

Figure 7(a). Saliency Maps for Basic Shooting Scenario: Single View

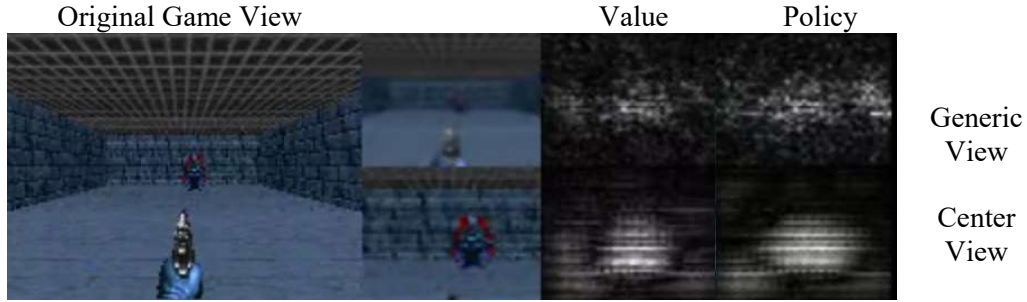
Figure 7(b). Saliency Maps for Basic Shooting Scenario: Dual View

Regarding Figure 7 (a, b), both maps are around the area where the monster is located. Generally, the maps of the single view agent look smoother than the dual view agent, due to the fact that the corresponding input frame of the single view agent is double the size of the dual view agent. Regarding dual view agents, the generic map is around the monster area, and the centered view is strongly focusing on the monster. This behavior of the two views is naturally and implicitly learned.

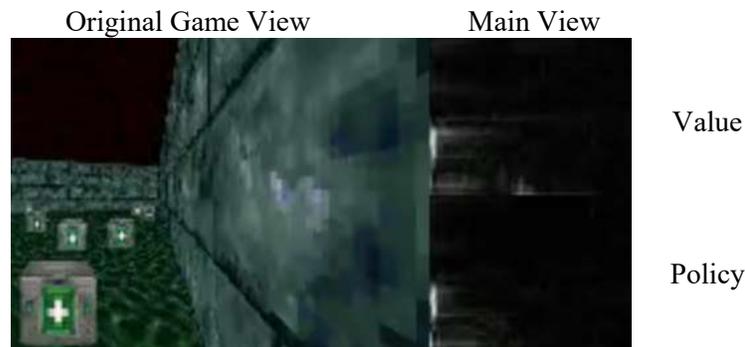
Figure 8(a). Saliency Maps for Health Gathering Scenario: Single View

In Figure 8(a), the network pays more attention to the lower left corner of the input frame, where there is no wall and there are medical kits. In Figure 8(b), both generic and center views attend around reasonable areas.

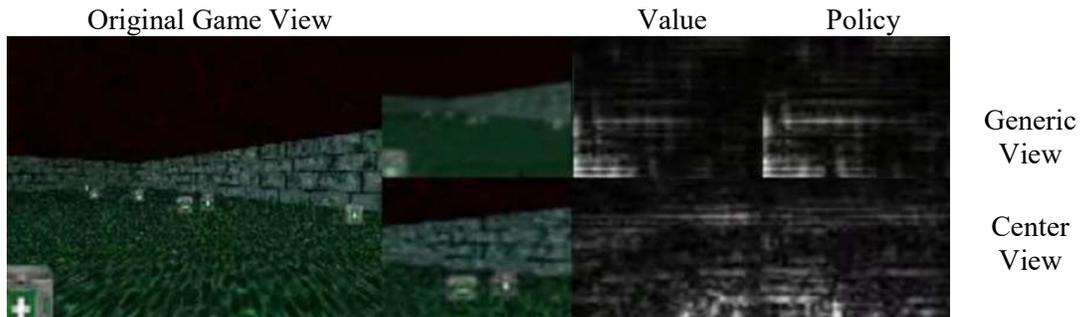

Figure 8(b). Saliency Maps for Health Gathering scenario: Dual View

For both testing game scenarios, at final training stages, the network learned to pay more attention to reasonable regions in the input frames for both value and policy estimations, without explicitly being trained to do so.

## 6. CONCLUSIONS AND FUTURE WORK

We introduce a modified architecture for speeding up the training process of robust deep reinforcement learning agents by dividing the input into dual views where the total input dimensionality is halved. The proposed architecture succeeds to have the same performance of A3C while achieving reduction in number of training parameters up to 30%. The dual view agent shows a robust performance when input frames are dropped at different probabilities. Visualization of saliency maps indicates reasonable agent learning. In conclusion, dividing the input is natural and suitable for a wide range of applications.

In future, we suggest to test the same architecture for training agents on different 3D environments, specifically for self-driving cars. TORCS [22] is considered a good realistic 3D environment for the task of self-driving. OpenAI Universe, which is introduced after OpenAI gym [23] contains many other environments. Finally, different settings for the dual view, such as direction and resolution, can be designed for specific task prior knowledge.